\documentclass{article}

\usepackage[nonatbib,final]{neurips_2022}
\usepackage{amsmath}
\usepackage{float}
\usepackage{graphicx}
\usepackage{pythonhighlight}
\PassOptionsToPackage{numbers}{natbib}

\usepackage[utf8]{inputenc} % allow utf-8 input
\usepackage[T1]{fontenc}    % use 8-bit T1 fonts
\usepackage{hyperref}       % hyperlinks
\usepackage{url}            % simple URL typesetting
\usepackage{booktabs}       % professional-quality tables
\usepackage{amsfonts}       % blackboard math symbols
\usepackage{nicefrac}       % compact symbols for 1/2, etc.
\usepackage{microtype}      % microtypography
\usepackage{xcolor}         % colors

\title{One-class systems seamlessly fit in the forward-forward algorithm}

\author{%
  Michael Hopwood\\
  Department of Statistics and Data Science\\
  University of Central Florida\\
  Orlando, FL\\
  \texttt{michael.hopwood@knights.ucf.edu} \\
}

\begin{document}

\maketitle

\begin{abstract}
The forward-forward algorithm \cite{hinton2022forward} presents a new method of training neural networks by updating weights during an inference, performing parameter updates for each layer individually.
This immediately reduces memory requirements during training and may lead to many more benefits, like seamless online training.
This method relies on a loss ("goodness") function that can be evaluated on the activations of each layer, of which can have a varied parameter size, depending on the hyperparamaterization of the network.
In the seminal paper, a goodness function was proposed to fill this need; however, if placed in a one-class problem context, one need not pioneer a new loss because these functions can innately handle dynamic network sizes. In this paper, we investigate the performance of deep one-class objective functions when trained in a forward-forward fashion.
The code is available at \url{https://github.com/MichaelHopwood/ForwardForwardOneclass}.
\end{abstract}

\section{Introduction}

The Forward-Forward algorithm \cite{hinton2022forward} is a new learning procedure for neural networks that updates network parameters immediately after the forward pass of a layer.
An objective (aka, "goodness") function is evaluated on the layer's latent output representations $G(h^{[l]} | \mathcal{I})$ conditioned upon some data integrity $\mathcal{I}$.
Integrity is broken down into positive and negative data; positive data is often thought of as correct data while negative data is incorrect data.
When positive data is passed into the model, weights that support the data (aka, neurons that fire with large weights) are awarded.
The assignment of these positive and negative data is subject to creativity with one of the most common practices being placing incorrect class assignments in the negative data. 

In a one-class problem context, it is assumed that the majority of the training dataset consists of “normal” data, and the model is assigned with determining the normality of the input data.
\textbf{Therefore, negative data is not required, and the objective function can be simplified to $G(h^{[l]})$.}
Many deep learning methods answer this anomaly detection problem via inspirations from support vector machines \cite{cortes1995support} like Deep SVDD \cite{ruff2018deep} or Deep OC-SVM \cite{sohn2020learning}.

\section{Methodology}

For a layer $l$ we compute a forward pass

$$h^{[l]} = \hbox{ReLU} \left( x W^{[l]} + b^{[l]} \right)$$

where $x \in \mathbb{R}^{n, p}$ is the data from the previous layer, $h \in \mathbb{R}^{n, q}$ is the transformed data, and $W^{[l]} \in \mathbb{R}^{p, q}$ and $b^{[l]} \in \mathbb{R}^{q}$ are the trained weights and biases. A forward pass of normal class data can be used to calculate the loss function at layer $l$ following some $G(h^{[l]})$. These $G(h^{[l]})$ can be any convex function; in the following table, we produce some candidate goodness functions.

\begin{table}[H]
    \centering
    \begin{tabular}{|c|l|l}
        \hline
        Method & Derivation\\ % & Gradients\\
        \hline
        Goodness & $\begin{aligned}
            \mathcal{L}(h^{[l]} ; \mathcal{W}) = \sum_{i=1}^N \sigma ( ||h^{[l]}||^2 - C )
        \end{aligned}$\\
        \hline
        GoodnessAdjusted & $\begin{aligned}\mathcal{L}(h^{[l]} ; \mathcal{W}) = \sum_{i=1}^N \log ( 1 + \exp( ||h||^2 - C ))\end{aligned}$ \\ 
        \hline
        HB-SVDD & $\begin{aligned}
                    \mathcal{L}(h^{[l]} ; \mathcal{W}) = \sum_{i=1}^N ||h^{[l]} - \mathbf{a}||^2                \end{aligned}$\\
                % &   \begin{aligned}
                %         \frac{\partial G}{\partial h^{[l]}} &= 2 h^{[l]} \quad \in \mathbb{R}^{n, q}\\
                %         \frac{\partial G}{\partial W^{[l]}} &= x^\top \frac{\partial G}{\partial h^{[l]}} \quad \in \mathbb{R}^{p, q} \\
                %         \frac{\partial G}{\partial b^{[l]}} &= \sum_{i \in n} \frac{\partial G}{\partial h_i^{[l]}}  \quad \in \mathbb{R}^{q}
                %         \end{aligned}\\
        \hline
        SVDD \cite{ruff2018deep} & $\begin{aligned}
                & \underset{\mathbf{a}, R , \xi_i}{\text{minimize}} \hspace{0.4cm} R^2 + C \sum_{i=1}^N \xi_i  \nonumber \\
                & \text{subject to} \hspace{0.3cm} ||h^{[l]} - \mathbf{a}||^2 \leq R^2 + \xi_i, \hspace{0.5cm} i=1,2,...,N \label{eq:1} \\
                & \text{and} \hspace{1.3cm} \xi_i \ge 0, \hspace{2.2cm} i =1,2,...,N\\
                \cline{1-2}
                & \mathcal{L}(h^{[l]}; R, \mathcal{W}) = R^2 + C \sum_{i=1}^N \max \bigl( 0, ||h^{[l]} - \mathbf{a}||^2 - R^2 \bigr) \end{aligned}$ \\
            % & \begin{aligned}
            %     \frac{\partial G}{\partial h^{[l]}} &= \begin{cases}
            %     2 h^{[l]} & \text{if } \quad ||h^{[l]} - \mathbf{a}||^2 - R^2 > 0 \\
            %     0 & \text{otherwise}
            %     \end{cases}
            % \end{aligned}\\
        \hline
        LS-SVDD     &$\begin{aligned}
                    & \underset{R, \mathbf{a}, \xi_i}{\text{minimize}} \hspace{0.4cm} R^2 + \frac{C}{2} \sum_{i = 1}^N \xi_i^2  \nonumber \\
                    & \text{subject to} \hspace{0.3cm} ||h^{[l]} - \mathbf{a}|| = R^2 + \xi_i, \hspace{0.5cm} i=1,2,...,N\\
                    \cline{1-2}
                    & \mathcal{L}(h^{[l]}; R, \mathcal{W}) = R^2 + \frac{C}{2} \sum_{i=1}^N \Bigl( 
                     ||h^{[l]} - \mathbf{a}||^2 - R^2 \Bigr)^2        \end{aligned}$\\
        % \hline
        % OCSVM   & \begin{aligned}
        %         & \underset{\mathbf{w}, \rho , \xi_i}{\text{minimize}} \hspace{0.4cm} \frac{1}{2} \mathbf{w}^T \mathbf{w} - \rho +  C \sum_{i = 1}^N \xi_i  \nonumber \\
        %         & \text{subject to} \hspace{0.3cm} \mathbf{w}^T h^{[l]} \ge \rho - \xi_i, \hspace{0.5cm} i=1,2,...,N \label{eq:1} \\
        %         & \text{and} \hspace{1.3cm} \xi_i \ge 0, \hspace{2.2cm} i =1,2,...,N \\
        %         \cline{1-2}
        %         & \mathcal{L}(h^{[l]}; \rho , \mathcal{W}) = \frac{1}{2} \mathbf{w}^T \mathbf{w} - \rho + C \sum_{i=1}^N \max \left(0, \rho - \mathbf{w}^T h^{[l]} \right)
        %         \end{aligned} \\
    \hline
    \end{tabular}
    \caption{Derivations of deep learning one-class "goodness" functions. Note that $\mathbf{a} = \frac{1}{N} \sum_{i=1}^N h^{[l]}_{i,j}$.}
    \label{tab:methodology}
\end{table}

The network's weights are updated sequentially, where inputs $h^{[l-1]}$ are passed through the layer to compute $h^{[l]}$, the loss $\mathcal{L}(h^{[l]})$ is calculated, and used to backpropagate using gradient descent 
\begin{align*}
W^{[l]} &= W^{[l]} + \frac{\lambda}{n} \frac{\partial G}{\partial W^{[l]}} \\
b^{[l]} &= b^{[l]} + \frac{\lambda}{n} \frac{\partial G}{\partial b^{[l]}}
\end{align*}

To convert the final embeddings $h^{[L]} \in \mathbb{R}^{n \times q}$ into an outlier probability, we pass them into the loss function to ascertain a distance value $D = \mathcal{L}(h^{[L]}) \in \mathbb{R}^{n}$ for each sample and then convert these distances to probabilities by normalizing by the maximum value, so $P = \frac{D}{\max(D)} \in \mathbb{R}^{n}$. In order to deem the sample an outlier, a threshold is deduced during training by evaluating $t = P_{(1-\nu)th \%}$. Therefore, an outlier is flagged via $I_{P>t}$. We utilize a $\nu=0.05$ for all settings. This method of ascertaining a threshold naturally reduces our chances of achieving 100\% accuracy, but it also reduces the chances of a type 2 error, which is important for outlier detection problems.

The code is written in PyTorch to leverage its built-in autodifferentiation tool. For the Forward-Forward implementation, gradients are computed at the end of each layer and the weights are updated according to the calculated autodifferentiated gradients and the optimizer. The normal backpropagation implmenetation conducts the weight update process for the weights in all layers after completing the forward pass on the last layer. So, while the forward-forward implementation has $L$ instantiated optimizers, the normal backpropagation method has 1 instantiated optimizer. For both cases, a stochastic gradient descent optimizer was used with no momentum and weight decay (see equations above). Early stopping is implemented by checking whether the backpropagation 

In order to make the experiments reproducible, random seeds were implemented. Across the 50 independent trials which were run for each parameter setting, a seed was $s=1...50$ was used when initializing the model parameters (e.g. weights and biases). For all independent trials, the same data split (e.g. train, valid, test) was used. This step is imperative, especially given the importance of the weight initialization for oneclass problem settings.

\subsection{Data}

The banknote authentication dataset \cite{Dua:2019} was used for evaluating the different methods. This data comprises images of both authentic and counterfeit banknotes captured using an industrial camera typically utilized for print inspection. The resulting images had a resolution of 400 x 400 pixels, and due to the object lens and distance to the subject, grayscale images with a resolution of approximately 660 dpi were obtained. The Wavelet Transform tool was employed to extract features from the images, resulting in 4 continuous features total, 3 features containing statistics of the Wavelet Transformed image (variance, skewness, kurtosis), and also the entropy of the image. The response variable is a binary value; 610 of the 1372 samples were deemed fake. 

\subsection{Evaluation}

This data was split into train, validation, and test splits. The training data trained the network weights. The validation data was used to decide early stopping. The test data was used to evaluate the model using accuracy, F1, and AUC. A grid search was conducted across the 5 loss functions (Table \ref{tab:methodology}), across 4 neural network architectures. Each setting was evaluated using 50 independent tests across different seeds, which impacted the network random initializations.

\section{Results}

\subsection{Forward Forward (FF) v. Normal Backpropagation (BP)}

The tabulated results are provided in Tables \ref{tab:results} \& \ref{tab:results_bp}.
The average accuracy for all experiments using BP was 57.6047\%; the FF experiments had an average value of 56.6287\%. Therefore, on average, BP experiments were 1\% more accurate. Similarly, BP was around 0.01 (i.e. 1\%) better in AUC with average BP and FF values of 0.549 and 0.538, Additionally, BP was around 0.025 (i.e. 2.5\%) better in the F1 score with average BP and FF values of 0.299 and 0.276. respectively. 
However, given the volatility of training deep oneclass models, it is worthwhile to compare the performance of the best models as opposed to the average model performance. Looking at all metrics, the best models achieve higher performance when trained using a FF pipeline; accuracy improves from 93.45\% to 94.18\%, F1 score improves from 0.9274 to 0.9375, and AUC improves from 0.9354 to 0.9461.

\subsection{Loss function evaluation}

In the forward forward evaluations, all of the best models used the goodness functions. They also perform well on average, with two of the three metrics having the highest average model performance when using them. Interestingly, the backpropagation evaluations all perform the best when using an LS-SVDD loss.

\section{Conclusion}

\begin{figure}[H]
    \centering
    {\Large \begin{flushleft} a)\end{flushleft}}\includegraphics[width=1.07\textwidth]{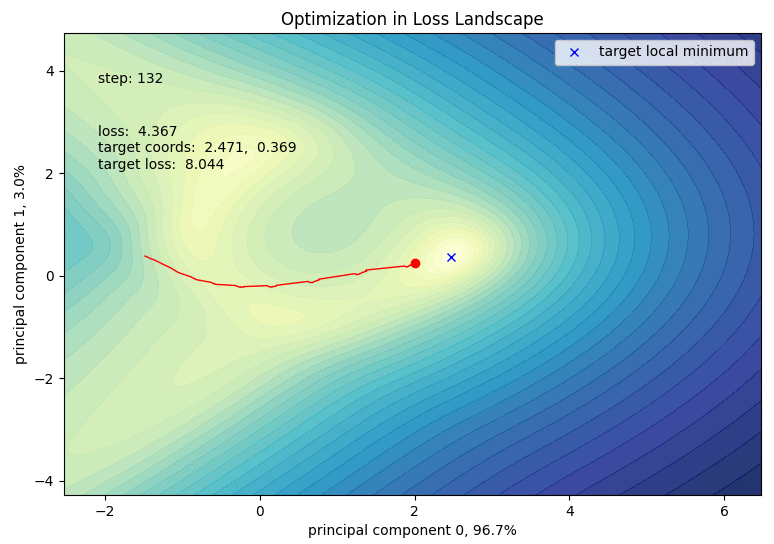}
    {\Large \begin{flushleft} b)\end{flushleft}}\includegraphics[width=1.07\textwidth]{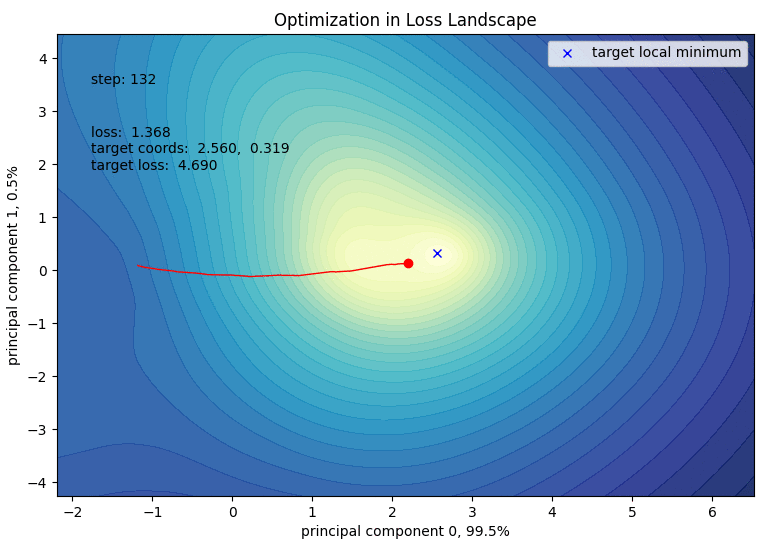}
    \caption{Loss landscapes for the neural network's a) first layer, and b) second layer when trained via forward-forward with a LS-SVDD loss.}
    \label{fig:my_label}
\end{figure}

In summary, the following conclusions were made:
\begin{enumerate}
    \item For one-class problems, forward-forward training shows comparable results to normal backpropagation in this case study (Table \ref{tab:results} and Table \ref{tab:results_bp})
    \item The goodness function is a viable loss candidate for one-class models (Table \ref{tab:results} and Table \ref{tab:results_bp})
    \item Forward-forward seemlessly enables the visualization of loss landscapes within the network, which can help gain insights into the learning process (Figure \ref{fig:my_label})
\end{enumerate}

Future work should be conducted to expand this study to deeper models and more benchmark data. Additionally, when training one-class problems using neural networks, many implementations find that pretraining the network weights using autoencoders are helpful, and sometimes, essential. Lastly, further work can introduce autoencoders into the training pipeline to regulate the model results across different random seeds.

\bibliographystyle{apalike}
\bibliography{refs}

\appendix
\section{Appendix}

\begin{table}[H]
\centering
\begin{tabular}{|c|cc|cc|cc|}
\hline
\textbf{} & \multicolumn{2}{c|}{\textbf{Accuracy (\%)}} & \multicolumn{2}{c|}{\textbf{F1}} & \multicolumn{2}{c|}{\textbf{AUC}} \\
\textbf{Method} & $\mu (\pm \sigma)$ & $\max$ & $\mu (\pm \sigma)$ & $\max$ & $\mu (\pm \sigma)$ & $\max$ \\ \hline
Goodness (4,10,10) & \multicolumn{1}{c|}{60.04 ($\pm$ 10.15 )} & 89.82 & \multicolumn{1}{c|}{0.2568 ($\pm$ 0.2681)} & 0.8923 & \multicolumn{1}{c|}{0.5598 ($\pm$ 0.115) } & 0.9035 \\
Goodness (4,25,25) & \multicolumn{1}{c|}{59.49 ($\pm$ 10.14 )} & \textbf{93.82} & \multicolumn{1}{c|}{0.2319 ($\pm$ 0.2686)} & \textbf{0.9333} & \multicolumn{1}{c|}{0.5529 ($\pm$ 0.1153) } & \textbf{0.942} \\
Goodness (4,50,50) & \multicolumn{1}{c|}{63.23 ($\pm$ 12.41 )} & 92.73 & \multicolumn{1}{c|}{0.3273 ($\pm$ 0.3099)} & 0.916 & \multicolumn{1}{c|}{0.5949 ($\pm$ 0.1392) } & 0.9238 \\
Goodness (4,100,100) & \multicolumn{1}{c|}{\textcolor{brown}{\textbf{65.21}} ($\pm$ 13.95 )} & 88.0 & \multicolumn{1}{c|}{\textbf{0.3704} ($\pm$ 0.3367)} & 0.8629 & \multicolumn{1}{c|}{\textcolor{brown}{\textbf{0.6177}} ($\pm$ 0.1567) } & 0.8764 \\
\hline
GoodnessAdjusted (4,10,10) & \multicolumn{1}{c|}{59.81 ($\pm$ 10.04 )} & 89.82 & \multicolumn{1}{c|}{0.2491 ($\pm$ 0.2639)} & 0.8923 & \multicolumn{1}{c|}{0.557 ($\pm$ 0.1136) } & 0.9035 \\
GoodnessAdjusted (4,25,25) & \multicolumn{1}{c|}{59.56 ($\pm$ 9.98 )} & \textcolor{brown}{\textbf{94.18}} & \multicolumn{1}{c|}{0.2384 ($\pm$ 0.2667)} & \textcolor{brown}{\textbf{0.9375}} & \multicolumn{1}{c|}{0.5541 ($\pm$ 0.1133) } & \textcolor{brown}{\textbf{0.9461}} \\
GoodnessAdjusted (4,50,50) & \multicolumn{1}{c|}{62.16 ($\pm$ 12.37 )} & 90.55 & \multicolumn{1}{c|}{0.3082 ($\pm$ 0.3035)} & 0.8879 & \multicolumn{1}{c|}{0.5836 ($\pm$ 0.1384) } & 0.8993 \\
GoodnessAdjusted (4,100,100) & \multicolumn{1}{c|}{\textbf{63.69} ($\pm$ 13.92 )} & 91.64 & \multicolumn{1}{c|}{\textbf{0.3372} ($\pm$ 0.3289)} & 0.9046 & \multicolumn{1}{c|}{\textbf{0.601} ($\pm$ 0.1554) } & 0.914 \\
\hline
HB-SVDD (4,10,10) & \multicolumn{1}{c|}{57.14 ($\pm$ 5.89 )} & 76.36 & \multicolumn{1}{c|}{0.1853 ($\pm$ 0.1757)} & 0.6829 & \multicolumn{1}{c|}{0.5261 ($\pm$ 0.0659) } & 0.7444 \\
HB-SVDD (4,25,25) & \multicolumn{1}{c|}{57.99 ($\pm$ 7.5 )} & 80.36 & \multicolumn{1}{c|}{0.2107 ($\pm$ 0.2154)} & 0.7523 & \multicolumn{1}{c|}{0.5363 ($\pm$ 0.0844) } & 0.7903 \\
HB-SVDD (4,50,50) & \multicolumn{1}{c|}{\textbf{60.6} ($\pm$ 9.05 )} & \textbf{86.18} & \multicolumn{1}{c|}{\textbf{0.298} ($\pm$ 0.2322)} & \textbf{0.8376} & \multicolumn{1}{c|}{\textbf{0.5669} ($\pm$ 0.101) } & \textbf{0.8559} \\
HB-SVDD (4,100,100) & \multicolumn{1}{c|}{58.29 ($\pm$ 8.27 )} & 80.73 & \multicolumn{1}{c|}{0.2322 ($\pm$ 0.227)} & 0.7558 & \multicolumn{1}{c|}{0.541 ($\pm$ 0.0919) } & 0.7936 \\
\hline
SVDD (4,10,10) & \multicolumn{1}{c|}{\textbf{48.2} ($\pm$ 5.4 )} & 61.09 & \multicolumn{1}{c|}{0.4169 ($\pm$ 0.2699)} & \textbf{0.6146} & \multicolumn{1}{c|}{0.4993 ($\pm$ 0.0167) } & 0.5615 \\
SVDD (4,25,25) & \multicolumn{1}{c|}{47.64 ($\pm$ 5.45 )} & 61.45 & \multicolumn{1}{c|}{0.4539 ($\pm$ 0.2527)} & \textbf{0.6146} & \multicolumn{1}{c|}{0.5004 ($\pm$ 0.0208) } & 0.5656 \\
SVDD (4,50,50) & \multicolumn{1}{c|}{46.21 ($\pm$ 4.47 )} & 60.0 & \multicolumn{1}{c|}{\textcolor{brown}{\textbf{0.5328}} ($\pm$ 0.1922)} & \textbf{0.6146} & \multicolumn{1}{c|}{0.5013 ($\pm$ 0.0139) } & 0.5509 \\
SVDD (4,100,100) & \multicolumn{1}{c|}{47.6 ($\pm$ 6.13 )} & \textbf{62.91} & \multicolumn{1}{c|}{0.5011 ($\pm$ 0.2094)} & \textbf{0.6146} & \multicolumn{1}{c|}{\textbf{0.5067} ($\pm$ 0.0234) } & \textbf{0.582} \\
\hline
LS-SVDD (4,10,10) & \multicolumn{1}{c|}{\textbf{54.97} ($\pm$ 5.15 )} & \textbf{71.64} & \multicolumn{1}{c|}{\textbf{0.138} ($\pm$ 0.1439)} & \textbf{0.6174} & \multicolumn{1}{c|}{\textbf{0.5032} ($\pm$ 0.0559) } & \textbf{0.6853} \\
LS-SVDD (4,25,25) & \multicolumn{1}{c|}{53.94 ($\pm$ 3.72 )} & 69.45 & \multicolumn{1}{c|}{0.1074 ($\pm$ 0.1137)} & 0.5484 & \multicolumn{1}{c|}{0.4917 ($\pm$ 0.0399) } & 0.6665 \\
LS-SVDD (4,50,50) & \multicolumn{1}{c|}{53.24 ($\pm$ 2.66 )} & 57.09 & \multicolumn{1}{c|}{0.0691 ($\pm$ 0.0692)} & 0.2561 & \multicolumn{1}{c|}{0.4828 ($\pm$ 0.024) } & 0.5206 \\
LS-SVDD (4,100,100) & \multicolumn{1}{c|}{53.56 ($\pm$ 2.13 )} & 57.45 & \multicolumn{1}{c|}{0.0537 ($\pm$ 0.0569)} & 0.183 & \multicolumn{1}{c|}{0.4845 ($\pm$ 0.0202) } & 0.523 \\
\hline
\end{tabular}
\caption{Results across 50 independent trials. \textbf{Forward forward}. Seed controlled.}
\label{tab:results}
\end{table}

\begin{table}[H]
\centering
\begin{tabular}{|c|cc|cc|cc|}
\hline
\textbf{} & \multicolumn{2}{c|}{\textbf{Accuracy (\%)}} & \multicolumn{2}{c|}{\textbf{F1}} & \multicolumn{2}{c|}{\textbf{AUC}} \\
\textbf{Method} & $\mu (\pm \sigma)$ & $\max$ & $\mu (\pm \sigma)$ & $\max$ & $\mu (\pm \sigma)$ & $\max$ \\ \hline
Goodness (4,10,10) & \multicolumn{1}{c|}{60.92 ($\pm$ 11.44 )} & 90.18 & \multicolumn{1}{c|}{0.2705 ($\pm$ 0.2917)} & 0.8898 & \multicolumn{1}{c|}{0.5695 ($\pm$ 0.1295) } & 0.901 \\
Goodness (4,25,25) & \multicolumn{1}{c|}{59.82 ($\pm$ 10.37 )} & \textbf{91.27} & \multicolumn{1}{c|}{0.2341 ($\pm$ 0.2771)} & \textbf{0.9062} & \multicolumn{1}{c|}{0.5564 ($\pm$ 0.1185) } & \textbf{0.9166} \\
Goodness (4,50,50) & \multicolumn{1}{c|}{62.97 ($\pm$ 12.2 )} & \textbf{91.27} & \multicolumn{1}{c|}{0.3219 ($\pm$ 0.3061)} & 0.8966 & \multicolumn{1}{c|}{0.5919 ($\pm$ 0.1367) } & 0.9066 \\
Goodness (4,100,100) & \multicolumn{1}{c|}{\textbf{65.22} ($\pm$ 13.97 )} & 88.0 & \multicolumn{1}{c|}{\textbf{0.3703} ($\pm$ 0.3371)} & 0.8629 & \multicolumn{1}{c|}{\textbf{0.6178} ($\pm$ 0.157) } & 0.8764 \\
\hline
GoodnessAdjusted (4,10,10) & \multicolumn{1}{c|}{61.08 ($\pm$ 11.53 )} & 90.18 & \multicolumn{1}{c|}{0.2742 ($\pm$ 0.2922)} & 0.8898 & \multicolumn{1}{c|}{0.5713 ($\pm$ 0.1305) } & 0.901 \\
GoodnessAdjusted (4,25,25) & \multicolumn{1}{c|}{59.87 ($\pm$ 10.17 )} & \textbf{90.91} & \multicolumn{1}{c|}{0.237 ($\pm$ 0.2737)} & \textbf{0.902} & \multicolumn{1}{c|}{0.5569 ($\pm$ 0.1162) } & \textbf{0.9125} \\
GoodnessAdjusted (4,50,50) & \multicolumn{1}{c|}{62.88 ($\pm$ 12.16 )} & \textbf{90.91} & \multicolumn{1}{c|}{0.3203 ($\pm$ 0.3048)} & 0.8918 & \multicolumn{1}{c|}{0.5909 ($\pm$ 0.1363) } & 0.9025 \\
GoodnessAdjusted (4,100,100) & \multicolumn{1}{c|}{\textcolor{brown}{\textbf{66.23}} ($\pm$ 14.11 )} & \textbf{90.91} & \multicolumn{1}{c|}{\textbf{0.3951} ($\pm$ 0.3409)} & 0.898 & \multicolumn{1}{c|}{\textcolor{brown}{\textbf{0.6292}} ($\pm$ 0.1586) } & 0.9083 \\
\hline
HB-SVDD (4,10,10) & \multicolumn{1}{c|}{57.43 ($\pm$ 5.89 )} & 78.18 & \multicolumn{1}{c|}{0.1987 ($\pm$ 0.1705)} & 0.717 & \multicolumn{1}{c|}{0.5295 ($\pm$ 0.0654) } & 0.7657 \\
HB-SVDD (4,25,25) & \multicolumn{1}{c|}{58.71 ($\pm$ 7.99 )} & 79.27 & \multicolumn{1}{c|}{0.236 ($\pm$ 0.2227)} & 0.6984 & \multicolumn{1}{c|}{0.5449 ($\pm$ 0.0894) } & 0.7672 \\
HB-SVDD (4,50,50) & \multicolumn{1}{c|}{\textbf{61.35} ($\pm$ 9.85 )} & \textbf{86.55} & \multicolumn{1}{c|}{\textbf{0.3238} ($\pm$ 0.2446)} & \textbf{0.8412} & \multicolumn{1}{c|}{\textbf{0.5762} ($\pm$ 0.1096) } & \textbf{0.8592} \\
HB-SVDD (4,100,100) & \multicolumn{1}{c|}{59.06 ($\pm$ 8.66 )} & 80.36 & \multicolumn{1}{c|}{0.2546 ($\pm$ 0.2337)} & 0.7453 & \multicolumn{1}{c|}{0.5497 ($\pm$ 0.0961) } & 0.7878 \\
\hline
SVDD (4,10,10) & \multicolumn{1}{c|}{47.43 ($\pm$ 5.01 )} & 60.73 & \multicolumn{1}{c|}{0.4472 ($\pm$ 0.2612)} & \textbf{0.6146} & \multicolumn{1}{c|}{0.4981 ($\pm$ 0.0173) } & 0.5574 \\
SVDD (4,25,25) & \multicolumn{1}{c|}{48.22 ($\pm$ 6.02 )} & 61.82 & \multicolumn{1}{c|}{0.4514 ($\pm$ 0.2462)} & \textbf{0.6146} & \multicolumn{1}{c|}{0.5041 ($\pm$ 0.0219) } & 0.5697 \\
SVDD (4,50,50) & \multicolumn{1}{c|}{\textbf{46.74} ($\pm$ 5.01 )} & 60.0 & \multicolumn{1}{c|}{\textcolor{brown}{\textbf{0.5128}} ($\pm$ 0.2091)} & \textbf{0.6146} & \multicolumn{1}{c|}{0.5023 ($\pm$ 0.0155) } & 0.5542 \\
SVDD (4,100,100) & \multicolumn{1}{c|}{48.49 ($\pm$ 6.68 )} & \textbf{63.27} & \multicolumn{1}{c|}{0.4737 ($\pm$ 0.2233)} & \textbf{0.6146} & \multicolumn{1}{c|}{\textbf{0.5092} ($\pm$ 0.0252) } & \textbf{0.5861} \\
\hline
LS-SVDD (4,10,10) & \multicolumn{1}{c|}{57.48 ($\pm$ 7.25 )} & 77.45 & \multicolumn{1}{c|}{\textbf{0.2126} ($\pm$ 0.2127)} & 0.7438 & \multicolumn{1}{c|}{0.5327 ($\pm$ 0.0818) } & 0.7708 \\
LS-SVDD (4,25,25) & \multicolumn{1}{c|}{\textbf{57.77} ($\pm$ 7.51 )} & \textcolor{brown}{\textbf{93.45}} & \multicolumn{1}{c|}{0.2064 ($\pm$ 0.1991)} & \textcolor{brown}{\textbf{0.9274}} & \multicolumn{1}{c|}{\textbf{0.5338} ($\pm$ 0.0842) } & \textcolor{brown}{\textbf{0.9354}} \\
LS-SVDD (4,50,50) & \multicolumn{1}{c|}{56.11 ($\pm$ 6.1 )} & 84.36 & \multicolumn{1}{c|}{0.1418 ($\pm$ 0.1737)} & 0.8201 & \multicolumn{1}{c|}{0.514 ($\pm$ 0.0689) } & 0.8395 \\
LS-SVDD (4,100,100) & \multicolumn{1}{c|}{54.33 ($\pm$ 4.15 )} & 72.0 & \multicolumn{1}{c|}{0.115 ($\pm$ 0.1118)} & 0.5838 & \multicolumn{1}{c|}{0.4955 ($\pm$ 0.0434) } & 0.6919 \\
\hline
\end{tabular}
\caption{Results across 50 independent trials. \textbf{Backpropagation}. Seed controlled.}
\label{tab:results_bp}
\end{table}

% \begin{figure}[H]
%     \centering
%     \includegraphics[width=1.2\textwidth]{figs/ff_bp_comparisons_accuracy_on_nndims.png}
%     \caption{Difference in performance of best performing (highest accuracy) models for each NN dimension setting and loss function.}
% \end{figure}

% \begin{figure}[H]
%     \centering
%     a) \includegraphics[width=0.8\textwidth]{figs/lossname_vs_accuracy.png}\\
%     b) \includegraphics[width=0.8\textwidth]{figs/lossname_vs_accuracy_bp.png}
%     \caption{Test accuracy across different loss functions. Each loss function was run through 8 network configurations, which were each run independently 50 times using different random seeds.}
% \end{figure}

% \begin{figure}[H]
%     \centering
%     \includegraphics{figs/hbsvd_41010_layer0_path.png}
%     \includegraphics{figs/hbsvd_41010_layer1_path.png}
%     \caption{Caption}
%     \label{fig:my_label}
% \end{figure}

\end{document}